\documentclass[letterpaper]{article} % DO NOT CHANGE THIS
\usepackage{aaai2027}  % DO NOT CHANGE THIS
\usepackage[hyphens]{url}  % DO NOT CHANGE THIS
\usepackage{graphicx} 
\graphicspath{{Figures/}{submission_folder/Figures/}}
\usepackage{natbib}  % DO NOT CHANGE THIS AND DO NOT ADD ANY OPTIONS TO IT
\usepackage{caption} % DO NOT CHANGE THIS AND DO NOT ADD ANY OPTIONS TO IT
\usepackage{algorithm}
\usepackage{algorithmic}

\usepackage{newfloat}
\usepackage{listings}
\DeclareCaptionStyle{ruled}{labelfont=normalfont,labelsep=colon,strut=off} % DO NOT CHANGE THIS
\floatstyle{ruled}
\newfloat{listing}{tb}{lst}{}
\floatname{listing}{Listing}

\usepackage{booktabs}

\title{What Not to Erase: Saliency-Constrained Augmentation for Low-Resource Handwritten Character Recognition}

\author {
    Paraskevi Platanou,
    John Pavlopoulos
}
\affiliations {
    Athens University of Economics and Business\\
    Archimedes, Athena Research Center, Greece\\
    \{platanou, annis\}@aueb.gr
}

\nocopyright

\begin{document}

\maketitle

\begin{abstract}
Optical Character Recognition (OCR) and handwritten character recognition (HCR) in low-resource historical material collections often rely on data augmentation, yet augmentation strategies are typically evaluated almost exclusively through predictive accuracy. This leaves open whether a model’s improved performance is accompanied by saliency stability. This question is particularly important in low-resource historical HCR, where recognition is performed on variable and partially degraded letterforms and two models with similar predictive performance may differ substantially in how consistently their saliency patterns respond to such variation. We therefore study
saliency-guided augmentation as an explanation-aware strategy for low-resource HCR. In this work, we study saliency-guided augmentation as an explanation-aware strategy for low-resource HCR. Rather than treating saliency maps only as post-hoc interpretations, we use them to define augmentation variants that perturb, preserve, or constrain visually informative regions of character images. We compare classical document-style transformations, random lacunae, and several saliency-constrained lacuna variants across multiple low-resource handwritten datasets and ImageNet-pretrained model families, including ResNet, Vision Transformers, and ConvNeXt-style architectures. 
Our experiments examine how different augmentation variants affect both recognition performance and explanation stability, showing that similar accuracy scores may correspond to substantially different attribution behavior. The results support a broader view of explainability: saliency can serve not only as a diagnostic tool after training, but also as a design and evaluation principle for robust augmentation in low-resource optical recognition. To evaluate the effect of augmentation beyond accuracy, we introduce the Explainability Consistency Index (ECI), a character-level metric that quantifies the stability of saliency patterns under controlled visual perturbations. 
We find that saliency-constrained augmentation can improve recognition over classical augmentation, but no single saliency strategy is consistently best. The preferred form of the constraint varies with both architecture and corpus, indicating that saliency is better treated as a mechanism for controlling augmentation than as a fixed augmentation
strategy.
In some settings, saliency-constrained training improves explanation consistency even when predictive performance is similar. This shows that predictive metrics alone can obscure differences in model sensitivity under perturbation, motivating explanation
consistency as a complementary evaluation dimension.
\end{abstract}


\section{Introduction}

Handwritten text recognition (HTR) in historical material collections remains a challenging problem for machine learning. Historical scripts exhibit substantial variation in writing style, material condition, ink density, stroke degradation, and visual fragmentation. At the same time, annotated training data are often scarce, since character- or line-level annotation requires expert knowledge and is expensive to produce. In such low-resource scenaria, data augmentation is a central strategy for improving generalization by artificially increasing the visual variability available during training.

Most augmentation strategies, however, are evaluated primarily through predictive performance metrics such as accuracy, macro-F1, or character error rate (CER). While these metrics are essential, they provide limited information about how a model reaches its decisions. This limitation is particularly important in low-resource optical recognition, where two models may obtain similar test scores while relying on different visual evidence. A model may perform well by using local features, background artifacts, or dataset-specific features rather than stable and morphologically meaningful features of the character shape.

In parallel, explainability methods for computer vision, including saliency maps, LIME, and Grad-CAM, have become widely used for inspecting model decisions. Yet such methods are most often applied post hoc, that is, they are used after training to visualize which input regions most influence a model's prediction, but they are rarely integrated into the design or evaluation of augmentation strategies themselves. This raises two questions: can explanation signals be used not only to interpret models, but also to guide and evaluate the augmentations used to train them, and can explanation
stability provide information about trained models that recognition metrics alone do not reveal?

We address this question in the scenario of low-resource handwritten character recognition (HCR). Character-level recognition provides a controlled example for studying how augmentation affects saliency stability before moving to full HTR pipelines. We propose an explainability-aware augmentation framework that compares classical document-style transformations, random lacunae, and several saliency-constrained lacuna variants. Rather than treating saliency as a simple target for erasure, we investigate how saliency can constrain where degradation is allowed to occur, including variants that perturb low-saliency regions, high-saliency regions, or protect high-saliency regions while applying foreground degradation.

To evaluate augmentation beyond predictive accuracy, we introduce the Explainability Consistency Index (ECI), a character-level metric that measures the stability of the visual evidence used by the model under controlled visual perturbations. ECI allows us to compare augmentation strategies not only by their recognition performance, but also by the consistency of the visual evidence used by the model. This is especially important in low-resource historical scripts, where robust recognition should depend on stable character morphology rather than accidental visual features.

We evaluate the proposed framework on two primary low-resource Greek character datasets and three pretrained model families: ResNet18, ViT-B/16, and ConvNeXt-Tiny. The primary datasets cover Hellenistic papyrological writing and Byzantine Greek manuscript writing, providing distinct material and visual domains. An extended ten-seed evaluation additionally introduces a Latin manuscript corpus as a cross-language robustness setting. The three-seed primary experiments show that saliency-constrained conditions can improve on classical augmentation, while the extended evaluation demonstrates that the magnitude and optimal form of this benefit depend on architecture and corpus. We therefore frame saliency not as a single universally best augmentation rule, but as a constraint space for controlling synthetic degradation in low-resource HCR.

Our contributions are summarized as follows:
\begin{itemize}
    \item We propose an explainability-aware augmentation framework for
    low-resource HCR, systematically comparing classical augmentation,
    random foreground lacunae, and saliency-constrained degradation
    policies that target, mix, or protect model-sensitive regions.

    \item We introduce the Explainability Consistency Index (ECI), combining
    top-region overlap and saliency correlation to characterize the stability
    of attribution patterns under controlled visual perturbations.

    \item We evaluate the framework across distinct historical writing
    domains and three pretrained architecture families, complemented by an
    extended ten-seed cross-language analysis that reveals substantial
    architecture- and corpus-dependence in the effect of saliency-guided
    augmentation.
\end{itemize}

\section{Related Work}
Explainability has become a central component of computer vision, offering tools for inspecting models that are otherwise often treated as black boxes. Explainability methods can be based on the following; attribution, activation, modified input and attention, according to recently adopted schemata \cite{cheng2025comprehensive}.

Attribution-based approaches attribute importance to input features by searching how changes in the input affect the output, backpropagating gradients to identify influential to the prediction pixels. Grad-CAM (Gradient-weighted Class Activation Mapping), which serves as an example of this method, uses the model's gradients to produce a heatmap highlighting image regions most relevant to the predicted class \cite{selvaraju2017gradcam}. Grad-CAM is class-discriminative and broadly applicable to CNN-based architectures without architectural changes. Other gradient-based methods include variants like Grad-CAM++ and SmoothGrad, which refine saliency maps by considering higher-order gradients or averaging over noisy inputs \cite{chattopadhyay2018gradcampp,smilkov2017smoothgrad}.

Activation- and feature-visualization methods inspect the internal representations learned by a neural network. Instead of explaining a prediction only through changes in the input or output, these methods examine intermediate feature maps, neuron activations, or gradients in order to identify which visual patterns activate different parts of the model. Early gradient-based saliency methods showed that the derivative of a class score with respect to the input image can be used to highlight pixels that influence a prediction \cite{simonyan2013deep}. Related visualization approaches, such as Deconvolutional Networks and guided backpropagation, project activation patterns back to the image space, revealing features learned at different layers of the network, from simple strokes and edges in early layers to more complex visual structures in deeper layers \cite{zeiler2014visualizing,springenberg2015striving}. Class Activation Mapping (CAM) is another influential approach, using the weights of a global average pooling layer to localize image regions that contribute to a class prediction \cite{zhou2016learning}. These methods are relevant to our work because they show that visual recognition models can be inspected not only through their final predictions, but also through the spatial evidence and internal representations that support those predictions.

In input-modification approaches the concept is that if a change applied to part of an input image causes significant change to the output, then the part change applied is significant. Classical examples include occlusion tests, with sliding patches across the image informing on prediction score drops \cite{zeiler2014visualizing}. Model-agnostic tools as LIME (Local Interpretable Model-Agnostic Explanations) randomly mask-out segments (e.g., superpixels) and fit a simple interpretable model to mimic the classifier locally \cite{ribeiro2016why}. The weights learned on various modified inputs indicate the most influential segments. Similarly, RISE (Randomized Input Sampling for Explanation) generates many random binary masks over the image and averages the output to build a saliency map of pixel importance \cite{petsiuk2018rise}. A major advantage of perturbation-based methods is that they can be model-agnostic, since they estimate importance by observing how predictions change under controlled input modifications. However, their faithfulness depends on the masking strategy, sampling procedure, and the extent to which the modified inputs remain realistic \cite{fong2017interpretable}.

Region-masking augmentation provides a related but distinct line of work. Cutout removes randomly positioned rectangular regions from training images \cite{devries2017cutout}, while Random Erasing randomly masks regions with varying size and aspect ratio \cite{zhong2020randomerasing}. GridMask instead removes image regions according to a structured grid pattern \cite{chen2020gridmask}.
These methods demonstrate that localized information removal can act as an effective regularizer, but the erased regions are not selected according to model-derived evidence. Our lacuna-based variants build on the general principle of localized degradation while restricting
perturbations to character foreground and, in the saliency-constrained conditions, using model sensitivity to determine where degradation is allowed or prevented.

With the rise of Vision Transformers (ViTs) and attention mechanisms, a new group of explainability methods refers to model attention weights for explanations. Vision Transformers internally compute attention scores indicating how much one patch influences another, and these scores can be visualized as heatmaps \cite{chefer2021transformer}. The appeal of attention-based explanations is that they offer a more global view since the model's receptive field is not local. However, attention is not an explanation per se, as research has already noted \cite{jain2019attention,wiegreffe2019attention}. Weights do not always align with a model's true reasoning and also transformers use feed-forward layers ignored by visualizations. 

Beyond post-hoc explanation, a growing number of scholars use explainability during model training and in particular to guide data augmentation and improve model generalization. One line of research focuses on saliency-based image augmentation for image classification. SaliencyMix \cite{uddin2021saliencymix}, SaliencyMix+ \cite{lee2025saliencymixplus}, and GuidedMixup
\cite{kang2023guidedmixup} use saliency to guide the selection or mixing of informative image regions. A closely related principle is introduced by KeepAugment \cite{gong2021keepaugment}, which protects informative regions while applying augmentation elsewhere in the image. These approaches show that model- or saliency-derived importance can inform not only post-hoc interpretation but also the construction of training examples. Saliency-guided augmentation has consequently been shown to improve upon unguided mixing or cropping strategies in
standard image-classification settings \cite{an2022saliencyguided}. Another direction is to incorporate human-annotated salient features into training data. In such tasks human experts annotate which parts of an image are truly important and then the training images are modified to emphasize these parts. Again, research has shown that images guided by (human) saliency achieved better accuracy and cross-dataset generalization than non-augmented images \cite{boyd2022humanaided}. Other effective strategies serve as a feedback loop in training, where explanations and augmentation feed into each other. ESSA (Explanation iterative Supervision via Saliency-guided data Augmentation) uses a repetitive cycle of steps; model training, saliency maps generation and saliency-based augmentation or modification \cite{gu2023essa}. Such explanation-guided learning approaches are still emerging and early results show they can lead to more robust models.

It is worth noting that most explainability-guided augmentation research has focused on general vision tasks such as natural image classification or object detection. In OCR and HTR, explainability has more often been used for model inspection than as a training-time principle for designing augmentation variants, especially in low-resource historical scripts. Explainability methods have been used mainly to analyze OCR models. However, using those explanations during training remains an open research area. Bringing explainability into OCR training could improve model reliability, much like in image classification tasks, ensuring that character recognition models learn to focus on the right features and not to rely on noise or artifacts. This can bridge the gap between explainability and high-performance OCR. 

Our work is related to both region-erasure augmentation and saliency-guided augmentation, but differs in its specific objective and constraints. Unlike generic masking methods such as Cutout, Random Erasing, and GridMask, lacuna placement is restricted to the estimated character foreground. Unlike patch-mixing approaches such as SaliencyMix and GuidedMixup, we do not transfer salient content between images. KeepAugment is closer to our protected variants in using importance information to avoid corrupting informative regions; our work extends this principle into a systematic study of saliency-controlled localized degradation for low-resource historical HCR, comparing protection with low-saliency, high-saliency, and mixed targeting policies under a specific character-foreground constraint and evaluating their effects on
both recognition and saliency stability.

\section{Method}

We propose an explainability-aware augmentation framework for low-resource handwritten character recognition. Given a training set $\mathcal{D}=\{(x_i,y_i)\}_{i=1}^N$, where $x_i$ is a character image and $y_i$ its class label, our goal is to train a classifier $f_\theta$ that is not only accurate, but also stable in the visual evidence it uses under controlled image modifications. The framework consists of three components: saliency extraction, saliency-constrained augmentation, and explanation-consistency evaluation.

\subsection{Saliency Extraction}

For each model family and random seed, we first train a source model using the original training images only. We then compute a character-level saliency map for each training image with respect to its ground-truth class. Let
\[
S_i = E(f_\theta, x_i, y_i)
\]
denote the saliency map produced by an explanation function $E$, where higher values indicate pixels that contribute more strongly to the model's prediction. In our experiments, $E$ is input-gradient saliency: we compute the absolute gradient of the target-class logit with respect to the input pixels and normalize the resulting map to $[0,1]$. This provides a local sensitivity estimate. In other words, pixels with larger gradient magnitude are those for which small input changes would most strongly affect the target-class score. We therefore use saliency as a measure of model sensitivity, not as a claim of human-interpretable attention. Unless otherwise stated, the target class is the ground-truth label for the training image. Saliency maps are normalized to the image domain and used to define candidate regions for augmentation. Since handwritten characters occupy only a subset of the image canvas, all saliency-based variants are constrained by a foreground mask $F_i$, obtained from the character strokes. This avoids selecting perturbation centers from background or padding regions. In all saliency-constrained augmentation variants, the saliency map is computed from a source classifier trained on the unaugmented training set. Classical augmentation and random foreground lacunae do not require saliency, while low-saliency, high-saliency, protected-random, and mixed lacunae use the source-model saliency map to define candidate or excluded regions.

\subsection{Augmentation Variants}

We compare classical augmentation, random lacunae, and saliency-constrained lacuna variants. Classical augmentation consists of small affine transformations, blur, brightness and contrast changes, and Gaussian noise. Lacuna-based augmentation simulates local degradation by masking small irregular regions of the character image. The twelve conditions and their design differences are summarized in Table~\ref{tab:augmentation_variants}.

Random lacunae sample degradation centers from the foreground mask without using saliency. Low-saliency lacunae perturb foreground regions that receive low saliency, while high-saliency lacunae perturb regions that receive high saliency. In contrast, saliency-protected random lacunae use saliency to define what should not be degraded: lacunae are sampled randomly from the foreground while excluding high-saliency regions. Formally, for image $x_i$, the allowed region for saliency-protected random lacunae is
\[
A_i = F_i \setminus H_i,
\]
where $F_i$ is the foreground mask and $H_i$ is the set of high-saliency pixels. This preserves the diversity of foreground degradation while protecting highly discriminative character regions.

\begin{table*}[t]
\centering
\small
\setlength{\tabcolsep}{5pt}
\begin{tabular}{llll}
\toprule
ID & Classical aug. & Lacuna placement & Saliency role \\
\midrule
A0  & No  & None & None \\
A1  & Yes & None & None \\
A2  & No  & Low-saliency foreground & Target low-saliency regions \\
A3  & No  & Random foreground & None \\
A4  & Yes & Low-saliency foreground & Target low-saliency regions \\
A5  & Yes & Random foreground & None \\
A6  & No  & High-saliency foreground & Target high-saliency regions \\
A7  & Yes & High-saliency foreground & Target high-saliency regions \\
A8  & Yes & Random foreground & Protect high-saliency regions \\
A9  & Yes & Mixed low/high saliency & Guide target selection \\
A10 & No  & Random foreground & Protect high-saliency regions \\
A11 & No  & Mixed low/high saliency & Guide target selection \\
\bottomrule
\end{tabular}
\caption{Augmentation conditions A0--A11. Classical augmentation
includes affine transformations, blur, brightness/contrast changes,
and Gaussian noise. All lacuna locations are restricted to the
character foreground.}
\label{tab:augmentation_variants}
\end{table*}

\begin{figure*}[t]
  \centering
  \includegraphics[width=\textwidth]{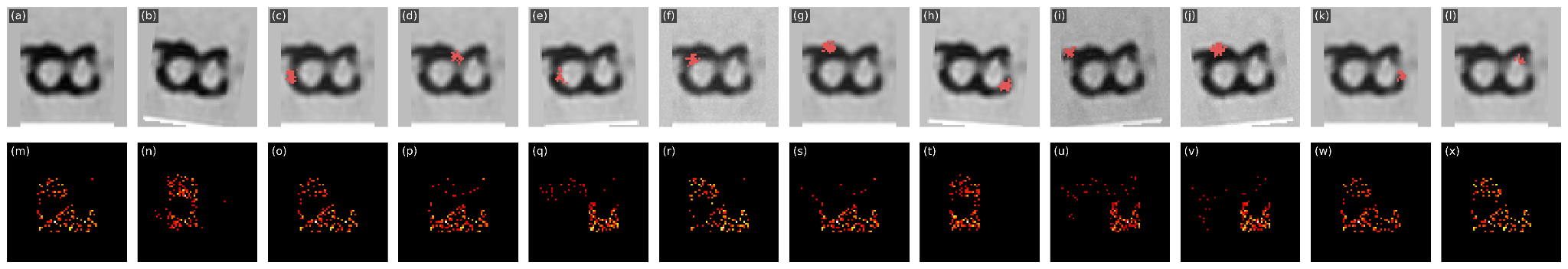}
  \caption{
  Augmentation variants and corresponding saliency visualizations for a representative handwritten character from HPGTR-Char. All panels show the preprocessed $64\times64$ model input, obtained by resizing and background-value padding. Saliency-guided regions in this example are computed using a ViT-B/16 source model trained on unaugmented images. Panels (a)--(l) correspond to conditions A0--A11 as defined in Table~\ref{tab:augmentation_variants}, and panels (m)--(x) show the corresponding foreground-masked saliency maps.
  }
\label{fig:augmentation_variants}
\end{figure*}

Figure~\ref{fig:augmentation_variants} illustrates the full augmentation family on a representative character image. Panels (a)--(l) show the resulting image variants, while panels (m)--(x) show the corresponding foreground-masked saliency maps. The visualization highlights the difference between variants that perturb foreground regions, variants that directly target salient regions, and variants that use saliency as a constraint to protect discriminative evidence. Besides showing the visible image perturbations, the figure also shows how the corresponding saliency maps change across augmentation variants. This qualitative comparison motivates the use of ECI: two augmented views may appear similarly plausible at the image level, but can induce different saliency patterns (e.g., panels (b), (f), and (j) versus their saliency maps in panels (n), (r), and (v)), while other perturbations may preserve the main regions of model sensitivity (e.g., panels (m), (o), (w), and (x)).

\subsection{Explainability Consistency Index}

Predictive performance alone does not indicate whether a model produces stable attribution patterns under controlled input changes. We therefore introduce the Explainability Consistency Index (ECI), which measures the agreement between saliency maps before and after a controlled perturbation. Let $x_i$ be an original image and $\tilde{x}_i=T(x_i)$ its perturbed version under transformation $T$. We compute saliency maps $S_i$ and $\tilde{S}_i$ for the original and perturbed images, respectively. To ensure that consistency is measured only over comparable regions, saliency maps are spatially aligned and evaluated over a valid region $R_i$ that excludes lacuna pixels and non-character background.

We report two complementary ECI scores. The first measures correlation between saliency values:
\[
\mathrm{ECI}_{corr}(x_i,T) =
\rho\left(
\mathrm{vec}(S_i \odot R_i),
\mathrm{vec}(\tilde{S}_i \odot R_i)
\right),
\]
where $\rho$ is the Pearson correlation coefficient, $\odot$ denotes element-wise masking, and $\mathrm{vec}(\cdot)$ vectorizes the valid pixels. The second measures overlap between the most salient regions:
\[
\mathrm{ECI}_{top}(x_i,T) =
\frac{
|Q_k(S_i,R_i) \cap Q_k(\tilde{S}_i,R_i)|
}{
|Q_k(S_i,R_i) \cup Q_k(\tilde{S}_i,R_i)|
},
\]
where $Q_k(S,R)$ denotes the top-$k$ salient pixels within region $R$. Dataset-level ECI is computed by averaging these scores over the evaluated images and perturbation views. Because $R_i$ excludes the lacuna region, in the present study, ECI is used as a measure of saliency-pattern stability over shared visible evidence.

\begin{figure*}[t]
  \centering
  \includegraphics[width=\textwidth]{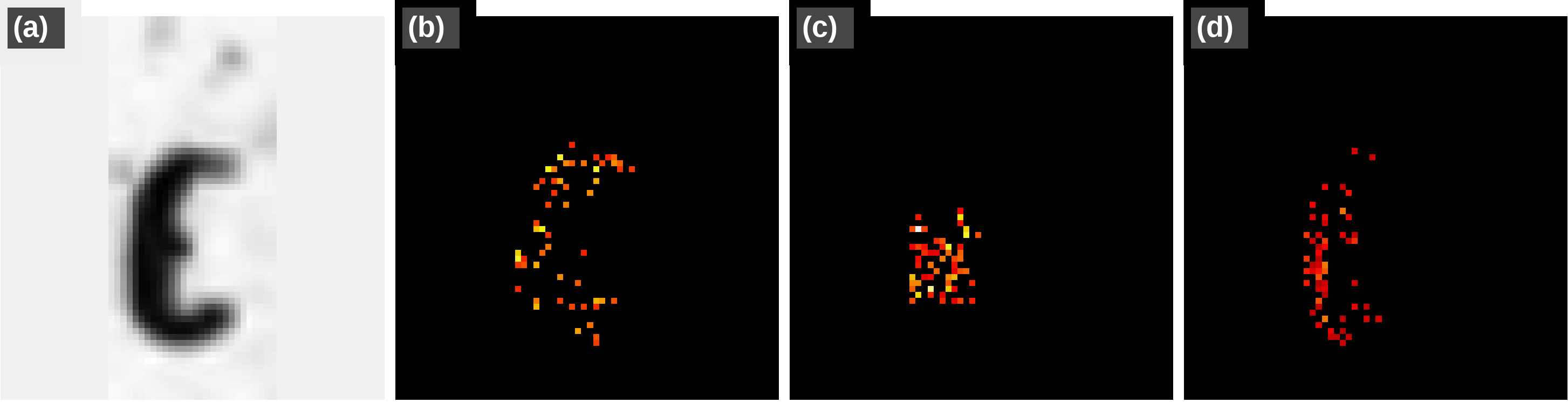}
  \caption{
  Architecture-specific saliency for the same handwritten character image from the HPGTR-Char dataset. 
  Panel (a) shows the original preprocessed $64\times64$ model input. Panels (b)--(d) show foreground-masked class-target saliency maps for ResNet18, ViT-B/16, and ConvNeXt-Tiny, respectively. 
  The example illustrates that different architectures can assign sensitivity to different regions of the same character image, motivating the use of explanation-consistency metrics in addition to predictive accuracy.
  }
  \label{fig:architecture_saliency}
\end{figure*}

ECI is not intended to replace accuracy. Instead, it complements predictive metrics by revealing whether augmentation strategies lead to stable or unstable attribution behavior. This distinction is important in low-resource handwritten recognition, where models with similar accuracy may rely on different, and potentially less generalizable, visual features. Figure~\ref{fig:architecture_saliency} provides a qualitative example of this effect. For the same input character, different architectures assign saliency to different regions of the character foreground. This motivates evaluating not only whether an augmentation improves recognition, but also if it promotes stable visual evidence.

\section{Experimental Setup}
\label{sec:experimental_setup}
Hyperparameters were fixed before the final multi-seed evaluation rather than selected through test-set optimization. Architecture-specific learning rates and batch sizes were chosen from standard fine-tuning configurations subject to GPU-memory constraints. The lacuna radius was fixed to $(1,2)$ after preliminary development experiments, and no test-set result was used for hyperparameter selection.

Experiments were conducted in Google Colab using NVIDIA A100 and L4 GPUs. The software environment used Python 3.12.13, PyTorch 2.11.0+cu128, torchvision 0.26.0+cu128, CUDA 12.8, scikit-learn 1.6.1, NumPy 2.0.2, Pillow 11.3.0, and OpenCV 4.13.0. GPU type and software versions were recorded for each experimental run.

\begin{algorithm}[t]
\caption{Saliency-Constrained Augmentation Pipeline}
\begin{algorithmic}[1]
\STATE Train an A0 source model for each architecture and seed.
\STATE Compute foreground-constrained input-gradient saliency maps.
\STATE Construct low-, high-, mixed-, and protected-saliency regions.
\FOR{each augmentation condition A0--A11}
    \STATE Train a new classifier using the corresponding transformation.
    \STATE Select the best epoch using validation macro-F1.
\ENDFOR
\STATE Select the augmentation condition using mean validation macro-F1.
\STATE Evaluate the selected condition once on the test set.
\STATE Compute ECI for explanation-consistency analysis.
\end{algorithmic}
\end{algorithm}

\subsection{Datasets}

The primary experiments use two character-level datasets of historical Greek
handwriting: Hell-Char \cite{pavlopoulos2027diachronic} and HPGTR-Char
\cite{platanou2026hpgtrchar}. They represent two distinct historical writing
domains: Hellenistic Greek papyri and Byzantine Greek manuscripts,
respectively. The extended ten-seed analysis additionally uses Corpus
Litterarum \cite{stroebel2025corpus}, a Latin manuscript dataset, as a
cross-language robustness setting.

\paragraph{Hell-Char.}
Hell-Char was introduced for diachronic representation learning of Ancient
Greek letterforms \cite{pavlopoulos2027diachronic} and is based on
character-level material originating from the Hell-Date collection
\cite{ferretti2025helldate}. After removing unknown labels, the version used
here contains 13,014 character images from 24 classes. We create a
group-aware split using the TM papyrus identifier, ensuring that characters
originating from the same papyrus cannot occur in more than one partition.
The resulting full split contains 8,232 training-pool images, 2,139
validation images, and 2,643 test images. To simulate a controlled
low-resource setting, we sample 10 training images per class from the
training pool, yielding 240 original training images, while leaving the
validation and test partitions unchanged.

\paragraph{HPGTR-Char.}
HPGTR-Char \cite{platanou2026hpgtrchar} contains 2,292 manually cropped
handwritten Greek character images derived from manuscript pages represented
in the HPGTR resource \cite{platanou2022hpgtr}. The character collection was
originally created and used in the Master's thesis on OCR for Greek
palaeographic manuscripts \cite{platanou2021ocr}. Each character filename
preserves positional provenance by encoding the source page, text-line
number, and character position within the line. For the present experiments,
we retain 1,996 records with available image files. Classes with
fewer than five samples are excluded, removing $\zeta$ and $\psi$. From the
remaining 22 classes, we construct a balanced low-resource split with 10
training, one validation, and one test image per class, yielding 220
training, 22 validation, and 22 test images.

All validation and test images remain unmodified in every experiment.
Dataset statistics are summarized in Table~\ref{tab:dataset_stats}.

\begin{table}[t]
\centering
\small
\begin{tabular}{lrrrrr}
\toprule
Dataset & Classes & Available & LR Train & Val & Test \\
\midrule
Hell-Char  & 24 & 13014 & 240 & 2139 & 2643 \\
HPGTR-Char & 22 & 2292  & 220 & 22   & 22 \\
\bottomrule
\end{tabular}
\caption{
Dataset statistics. ``Available'' denotes the character images
available in the corresponding dataset release before the low-resource experimental subsampling used here. For HPGTR-Char, 1,996 alphabetic records
with available image files remain after the initial eligibility filtering; the final experiment retains 22 sufficiently supported classes. Hell-Char
uses a document-grouped split by TM papyrus identifier, whereas HPGTR-Char uses a balanced low-resource split. Validation and test images are never augmented.}
\label{tab:dataset_stats}
\end{table}

\paragraph{Data availability.}
Hell-Char is released with the resources accompanying
\cite{pavlopoulos2027diachronic}.\footnote{\url{https://github.com/ipavlopoulos/diachronic-greek-letterforms}}
HPGTR-Char is publicly available;\footnote{\url{https://github.com/vivianpl/HPGTR-Char}} the corresponding source
manuscript-page resource is HPGTR
\cite{platanou2022hpgtr}.
Corpus Litterarum is publicly available through Zenodo
\cite{stroebel2025corpus}.

\subsection{Preprocessing}

All images are converted to grayscale. We estimate a foreground mask using Otsu thresholding, crop the character foreground with a small margin, and map the crop to a fixed $64\times64$ canvas using aspect-ratio-preserving resizing. Padding is filled with an estimated background value computed from border pixels. The same preprocessing is used for all datasets, architectures, and augmentation conditions. For ImageNet-pretrained backbones, the $64\times64$ grayscale tensor is repeated to three channels, upsampled to $224\times224$, and normalized.

\subsection{Augmentation Protocol}

We compare twelve training conditions, A0--A11. A0 uses only the original training images. A1 applies classical augmentation. A2, A3, A6, A10, and A11 are lacuna-only variants, while A4, A5, A7, A8, and A9 combine lacuna placement with classical augmentation.

For augmented conditions, each original training image contributes ten views per epoch: one original-equivalent view and nine stochastic augmented views. Thus, the effective number of training views per epoch is 2,400 for Hell-Char and 2,200 for HPGTR-Char. A0 uses 240 and 220 original training images per epoch, respectively. Validation and test images are never augmented.

Classical augmentation consists of small affine transformations, Gaussian blur, brightness and contrast jitter, and Gaussian noise. Lacunae are applied with probability 1.0 for lacuna-based variants, using radius range $(1,2)$. Saliency-constrained variants use saliency maps extracted from an A0 source model trained with the same architecture and random seed. Protected-random variants exclude high-saliency regions from the candidate foreground region.

\subsection{Architectures and Training}

We evaluate three architecture families: ImageNet-pretrained ResNet18, ImageNet-pretrained ViT-B/16, and ImageNet-pretrained ConvNeXt-Tiny.

All models are trained with cross-entropy loss and the AdamW optimizer with weight decay $10^{-4}$. Checkpoints are selected according to validation macro-F1. Early stopping uses a patience of five epochs. Experiments are repeated across three random seeds: 13, 42, and 123 (more seeds in the Appendix). For each run, the corresponding seed is applied to Python's
\texttt{random} module, NumPy, PyTorch CPU operations, and all CUDA
devices. Dataset splits are generated once using a fixed split seed
and stored for reuse across architectures and augmentation conditions. Data-loader generators and workers are seeded deterministically. Architecture-specific learning rates, maximum epochs, and batch sizes are shown in Table~\ref{tab:training_details}.

\begin{table}[t]
\centering
\small
\begin{tabular}{lrrrr}
\toprule
Architecture & LR & Epochs & Batch & Initialization \\
\midrule
ResNet18 & $10^{-4}$ & 20 & 16 & ImageNet \\
ViT-B/16 & $5{\times}10^{-5}$ & 15 & 8 & ImageNet \\
ConvNeXt-Tiny & $5{\times}10^{-5}$ & 15 & 8 & ImageNet \\
\bottomrule
\end{tabular}
\caption{
Architecture-specific training settings. All models trained with AdamW, cross-entropy loss, weight decay $10^{-4}$, early stopping patience (5 epochs), checkpoint selection by val. macro-F1. Experiments repeated (seeds 13, 42, 123).
}
\label{tab:training_details}
\end{table}

\subsection{Evaluation Metrics}
We report accuracy, macro-F1, and balanced accuracy. Macro-F1 is treated as the primary recognition metric because the experiments are low-resource and include class imbalance, especially in Hell-Char. For the primary metric, macro-F1, we report mean and standard error of the mean (SEM) over three random seeds. Accuracy measures the proportion of correctly classified test images. Balanced accuracy is the unweighted mean of per-class recall and is
reported to reduce the influence of class-frequency imbalance. Macro-F1 is the unweighted mean of class-specific F1 scores and serves as the primary recognition metric. We report variability across seeds using the standard error of the mean, computed as the sample standard deviation divided by the square root of the number of seeds. We further run exploratory Friedman tests across augmentation variants and paired Wilcoxon signed-rank tests with Holm correction for selected comparisons; given the small number of seeds, these tests are interpreted descriptively rather than as definitive evidence of significance. 

For explanation consistency, we compute ECI on the training split using up to 10 images per class and five perturbation views per image. Saliency agreement is evaluated over foreground-aligned regions, excluding lacuna pixels and non-character background. We report both top-region overlap, using the top 15\% most salient pixels within the valid region, and saliency correlation.

\section{Results}
\label{sec:results}

\subsection{Character Recognition Performance}
Table~\ref{tab:best_results} reports the best-performing augmentation condition for each dataset and architecture, using macro-F1 as the primary metric. Across the six primary dataset--architecture settings, a saliency-constrained augmentation variant obtains the highest mean macro-F1 under the three-seed protocol. Because architecture-specific results within each dataset share the same underlying split, we treat this pattern descriptively rather than as six independent statistical replications. The extended ten-seed analysis in the supplementary material provides a stricter assessment of robustness and shows that the magnitude and optimal form of the saliency benefit are architecture- and corpus-dependent.

At the level of individual augmentation variants, Friedman and Holm-corrected paired Wilcoxon tests over three random seeds did not yield statistically significant pairwise differences. We therefore interpret the per-variant comparisons as directional evidence rather than definitive pairwise significance.

\begin{table*}[t]
\centering
\small
\renewcommand{\arraystretch}{.6} 
\setlength{\tabcolsep}{8pt}\begin{tabular}{llrrrrrrr}
\toprule
Dataset & Architecture & Best & Macro-F1 $\pm$ SEM & Acc. & Bal. Acc. & $\Delta$A1 & ECI$_{top}$ & ECI$_{corr}$ \\
\midrule
Hell-Char & ResNet18 & A8 & $0.564 \pm 0.011$ & 0.586 & 0.601 & +0.024 & 0.182 & 0.283 \\
Hell-Char & ViT-B/16 & A9 & $0.609 \pm 0.004$ & 0.629 & 0.639 & +0.008 & 0.190 & 0.288 \\
Hell-Char & ConvNeXt-Tiny & A8 & $0.634 \pm 0.002$ & 0.653 & 0.673 & +0.026 & 0.276 & 0.475 \\
\midrule
HPGTR-Char & ResNet18 & A4 & $0.687 \pm 0.018$ & 0.758 & 0.758 & +0.061 & 0.157 & 0.217 \\
HPGTR-Char & ViT-B/16 & A9 & $0.821 \pm 0.003$ & 0.864 & 0.864 & +0.139 & 0.176 & 0.262 \\
HPGTR-Char & ConvNeXt-Tiny & A9 & $0.770 \pm 0.003$ & 0.818 & 0.818 & +0.018 & 0.266 & 0.446 \\
\bottomrule
\end{tabular}
\caption{
Best augmentation condition for each dataset and architecture, selected by mean macro-F1 over three random seeds. Macro-F1 is reported as mean $\pm$ standard error of the mean (SEM). $\Delta$A1 denotes the macro-F1 difference between the best method and classical augmentation. ECI scores are reported for the selected method. A saliency-constrained condition obtains the highest mean macro-F1 in all six primary dataset--architecture settings, although no Holm-corrected pairwise comparison reaches significance with three seeds.
}
\label{tab:best_results}
\end{table*}

On Hell-Char, the validation-selected saliency-constrained conditions obtain higher mean macro-F1 than classical augmentation for all three pretrained architectures. A8, which combines classical augmentation with saliency-protected random lacunae, achieves the best macro-F1 for ResNet18 and ConvNeXt-Tiny, while A9, the mixed low/high-saliency condition, performs best for ViT-B/16.

On HPGTR-Char, saliency-constrained variants are also selected for all three architectures. A4 is best for ResNet18, while A9 is best for ViT-B/16 and ConvNeXt-Tiny. The improvement over classical augmentation is particularly large for ViT-B/16, where macro-F1 increases from 0.682 under A1 to 0.821 under A9.

\subsection{Explanation Consistency}

The ECI results clarify why predictive performance alone is insufficient for evaluating augmentation. By construction, A0 yields nearly perfect explanation consistency because no perturbation is applied. Similarly, several lacuna-only variants can produce high ECI values while still yielding substantially lower recognition performance. High explanation consistency alone is therefore not the objective: a useful augmentation should improve recognition while avoiding unnecessary instability in the visual evidence used by the model.

The strongest balance is observed for methods that combine classical augmentation with saliency-constrained lacuna placement. Across datasets and architectures, the classical-plus-saliency family achieves the highest average macro-F1 among augmentation families, while maintaining ECI values comparable to classical augmentation. In contrast, classical augmentation combined with random foreground lacunae is less effective, indicating that the location of synthetic degradation matters. The comparison between random, low-saliency, high-saliency, protected-random, and mixed variants supports the view that saliency is useful not simply as a target for erasure, but as a constraint for controlling where degradation should or should not occur. These findings align with the qualitative examples in Figures~\ref{fig:augmentation_variants} and~\ref{fig:architecture_saliency}.

\section{Discussion}
\label{sec:discussion}
The main implication of our experiments is not that one saliency rule is universally best, but that model-derived sensitivity provides a useful constraint space for controlling synthetic degradation. Directly targeting high-saliency regions, perturbing low-saliency regions, protecting high-saliency regions, and mixing the two targeting strategies lead to different outcomes across architectures and corpora. The selected condition therefore varies by setting, while classical and random-foreground augmentation remain strong controls. This architecture- and corpus-dependent
behaviour argues against treating saliency-guided augmentation as a single fixed recipe and instead motivates validation-based selection among saliency-constrained policies.

ECI complements the recognition results by making this balance visible. High explanation consistency alone is not sufficient, since uninformative or weak perturbations may preserve saliency without improving recognition. Conversely, accuracy gains may be less desirable if they correspond to unstable or incoherent saliency patterns. ECI should therefore be interpreted jointly with macro-F1: useful augmentation should improve recognition while preserving coherent visual evidence over comparable foreground regions.

Several limitations remain. The experiments focus on isolated character recognition rather than full line- or word-level HTR, where segmentation ambiguity, ligatures, context, and layout introduce additional challenges. The saliency constraints also depend on the quality of the A0 source model; if this model is uncertain or relies on non-generalizable features, its saliency maps may provide imperfect guidance. The primary experiments use three random seeds and therefore provide limited power for variant-level statistical comparisons; the extended analysis partly addresses this issue with ten matched seeds, but does not eliminate the small-test-set limitation of HPGTR-Char.

The current ECI analysis has two further limitations. First, it is computed on training images and therefore measures in-sample consistency rather than attribution stability on unseen examples. Second, because lacuna pixels are excluded from the shared comparison region, ECI measures the consistency of surviving visible evidence but does not penalize the removal of originally salient evidence. Future work will therefore evaluate trained conditions under a shared held-out perturbation bank and complement ECI with an explicit measure of removed saliency mass.

\section{Conclusion}
\label{sec:conclusion}

We introduced an explainability-aware augmentation framework for low-resource handwritten character recognition in which source-model saliency is used to control synthetic degradation rather than only as a post-hoc diagnostic. We also introduced ECI as a complementary measure of saliency-pattern stability under controlled perturbations. Across two primary Greek character datasets and three pretrained architecture families, saliency-constrained conditions are frequently competitive with or stronger than classical augmentation in the three-seed evaluation. An extended ten-seed analysis, including a Latin manuscript corpus, shows that the magnitude and optimal form of this benefit are architecture- and corpus-dependent. These findings support saliency as a useful constraint space for augmentation design rather than as a universally superior augmentation rule. Future work will evaluate explanation consistency on held-out perturbations, explicitly quantify removed salient evidence, and extend the framework from isolated characters to line- and word-level HTR.

\section*{Acknowledgments}
This work has been partially supported by project MIS 5154714 of the National Recovery and Re-silience Plan Greece 2.0 funded by the European Union under the NextGenerationEU Program.

\appendix
\section{Appendix}
\label{app:supplementary}

The experiments reported in this appendix provide an extended ten-seed robustness analysis of the primary study. The protocol uses ten random seeds, selects augmentation conditions exclusively by mean validation macro-F1, and applies paired seed-level statistical tests. In addition to Hell-Char and HPGTR-Char, it introduces Corpus Litterarum as a cross-language evaluation corpus, testing whether the observed augmentation behaviour transfers from Greek to Latin
letterforms.

\subsection{Datasets for the Extended Experiments}

The extended ten-seed evaluation uses Hell-Char, HPGTR-Char, and Corpus Litterarum. These datasets cover Greek papyri, Greek manuscripts, and Latin manuscripts. Corpus Litterarum is included specifically to test cross-language generalization, allowing us to examine whether the augmentation and explanation-consistency findings transfer from Greek to Latin letterforms rather than remaining specific to one script tradition.

Hell-Char uses the Hell-Char character annotations derived from Hell-Date and introduced by Pavlopoulos et al. (2026). After excluding unknown labels, it contains 13,014 images from 24 classes. We split by TM-papyrus identifier so that no document occurs in multiple partitions: 8,232 images form the training pool, 2,139 validation, and 2,643 test. Ten training images per class are sampled from the training pool, giving 240 low-resource training images; validation and test remain complete.

Corpus Litterarum is a public line-based dataset of eighth-century Latin manuscript writing with 2,152 line images, 44,407 character annotations, 73 labels, and predefined line-level splits (Ströbel 2025). Unlike the Greek datasets, it provides Latin letterforms and therefore serves as a cross-language and cross-tradition generalization test. We retain 20 alphabetic classes with sufficient support, preserve the official line partitions, and sample 4 training, 10 validation, and 20 test characters per class with a fixed sampling seed, giving 80/200/400 images.

HPGTR-Char is a Byzantine Greek dataset of Greek characters. The final 22-class subset contains 264 images, split into 10 training, one validation, and one test image per class (220/22/22).

\begin{table}[t]
\centering
\small
\setlength{\tabcolsep}{3.5pt}
\begin{tabular}{lrrrrr}
\toprule
Dataset & Classes & Shots & Train & Val & Test \\
\midrule
Hell-Char & 24 & 10 & 240 & 2139 & 2643 \\
HPGTR-Char & 22 & 10 & 220 & 22 & 22 \\
Corpus Litterarum & 20 & 4 & 80 & 200 & 400 \\
\bottomrule
\end{tabular}
\caption{Experimental splits before augmentation. Shots denotes original training images per class. Validation and test images are never augmented.}
\label{tab:app_dataset_stats}\end{table}

\subsection{Preprocessing and Augmentation}

Images are converted to grayscale; Otsu thresholding estimates the foreground, which is cropped with a small margin and mapped to a \(64\times64\) canvas by aspect-ratio-preserving resizing and border-median padding. For ImageNet-pretrained backbones, the tensor is repeated to three channels, resized to \(224\times224\), and normalized.

A0 uses one original view per image. A1--A11 use ten views per image per epoch: one original-equivalent and nine stochastic augmented views. Effective epoch sizes are therefore 2,400 for Hell-Char, 2,200 for HPGTR-Char, and 800 for Corpus Litterarum. Lacunae are applied with probability 1.0 in lacuna-based conditions. Validation and test transformations contain preprocessing only.

\subsection{Architectures and Training}

We fine-tune ImageNet-pretrained ResNet18, ViT-B/16, and ConvNeXt-Tiny with cross-entropy, AdamW, weight decay \(10^{-4}\), and early-stopping patience 5. Seeds are 13, 42, 123, 7, 19, 31, 67, 101, 211, and 509. Each seed is applied to Python, NumPy, PyTorch CPU/CUDA, data-loader generators, and workers. Splits are generated once with fixed split seeds and stored for reuse. Table~\ref{tab:training_details} lists architecture-specific settings.

\begin{table}[t]
\centering
\small
\setlength{\tabcolsep}{4pt}
\begin{tabular}{lrrr}
\toprule
Architecture & LR & Epochs & Batch \\
\midrule
ResNet18 & \(10^{-4}\) & 20 & 16 \\
ViT-B/16 & \(5{\times}10^{-5}\) & 15 & 8 \\
ConvNeXt-Tiny & \(5{\times}10^{-5}\) & 15 & 8 \\
\bottomrule
\end{tabular}
\caption{Training settings. All models use ImageNet initialization; checkpoints are selected by validation macro-F1.}
\label{tab:app_training_details}
\end{table}

\subsection{Evaluation and Statistical Analysis}

Macro-F1 is the primary recognition metric; accuracy and balanced accuracy are also recorded. We report means and standard errors over ten seeds. For each dataset--architecture pair, the final condition is selected only by mean validation macro-F1 across A0--A11; test metrics are then reported without test-based reselection. The validation-selected saliency condition is compared with A1 and A5 using an exact paired sign-flip test as the primary test, paired Wilcoxon as a supplementary test, 95\% confidence intervals, and rank-biserial effect sizes. Holm correction is applied within each confirmatory metric family; Friedman tests across A0--A11 are exploratory.

ECI is secondary and is not used for method selection. It is computed on up to 10 training images per class with five perturbation views per image, using the top 15\% salient pixels for \(\mathrm{ECI}_{top}\). All comparisons use matched seeds and exclude lacuna and non-character pixels from valid regions.

\subsection{Recognition Performance}

Table~\ref{tab:best_results} reports the condition selected by validation for each setting under the extended ten-seed, validation-only protocol. Saliency-informed variants are selected in eight of nine settings; Corpus Litterarum with ViT-B/16 selects A5. Hell-Char selects A8 for ResNet18 and ConvNeXt-Tiny and A9 for ViT-B/16. Their test macro-F1 gains over A1 are 0.009, 0.019, and 0.009, respectively; ConvNeXt A8 is effectively tied with A5.

HPGTR-Char selects A7, A4, and A9. The largest recognition trend is ResNet18 A7 over A1 (\(\Delta=0.034\)), but it is not significant after Holm correction (exact \(p_{\mathrm{Holm}}=0.305\)); no HPGTR-Char recognition comparison reaches corrected significance. On Corpus Litterarum, A7 is selected for ResNet18, A5 for ViT-B/16, and A8 for ConvNeXt-Tiny. No saliency condition significantly improves recognition over A1 or A5 after correction, confirming that classical and random-lacuna baselines remain competitive.

\begin{table*}[t]
\centering
\small
\setlength{\tabcolsep}{3.2pt}
\begin{tabular}{lllrrrrrr}
\toprule
Dataset & Architecture & Sel. & Test F1 $\pm$ SEM & Acc. & $\Delta$A1 & $\Delta$A5 & ECI$_{top}$ & ECI$_{corr}$ \\
\midrule
Hell-Char & ResNet18 & A8 & $0.554\pm0.007$ & 0.578 & +0.009 & +0.005 & 0.176 & 0.270 \\
Hell-Char & ViT-B/16 & A9 & $0.606\pm0.005$ & 0.623 & +0.019 & +0.004 & 0.188 & 0.283 \\
Hell-Char & ConvNeXt-Tiny & A8 & $0.622\pm0.006$ & 0.641 & +0.009 & -0.0002 & 0.276 & 0.475 \\
\midrule
HPGTR-Char & ResNet18 & A7 & $0.683\pm0.023$ & 0.750 & +0.034 & +0.007 & 0.159 & 0.219 \\
HPGTR-Char & ViT-B/16 & A4 & $0.736\pm0.012$ & 0.786 & +0.017 & +0.017 & 0.169 & 0.245 \\
HPGTR-Char & ConvNeXt-Tiny & A9 & $0.780\pm0.016$ & 0.827 & +0.005 & -0.004 & 0.269 & 0.452 \\
\midrule
Corpus Litterarum & ResNet18 & A7 & $0.829\pm0.006$ & 0.830 & +0.008 & -0.006 & 0.161 & 0.223 \\
Corpus Litterarum & ViT-B/16 & A5 & $0.852\pm0.007$ & 0.851 & +0.006 & 0.000 & 0.159 & 0.209 \\
Corpus Litterarum & ConvNeXt-Tiny & A8 & $0.866\pm0.007$ & 0.865 & +0.010 & +0.004 & 0.227 & 0.371 \\
\bottomrule
\end{tabular}
\caption{Validation-selected conditions evaluated on test data over ten seeds. $\Delta$A1 and $\Delta$A5 are test macro-F1 differences from classical augmentation and classical+random foreground lacunae. ECI is reported for the selected condition.}
\label{tab:app_best_results}
\end{table*}

\subsection{Explanation Consistency}

ECI must be interpreted jointly with recognition: A0 and weak lacuna-only conditions can preserve saliency while underperforming. HPGTR-Char shows no corrected ECI differences between its selected saliency variants and A1/A5. The clearest separation occurs on Corpus Litterarum with ConvNeXt-Tiny. A8 and A5 have similar test macro-F1 (0.866 vs. 0.862), but A8 has higher top-region consistency (\(\Delta=0.0078\), exact \(p_{\mathrm{Holm}}=0.0469\), rank-biserial \(=0.891\)) and saliency correlation (\(\Delta=0.0171\), \(p_{\mathrm{Holm}}=0.0117\), rank-biserial \(=1.0\)); the correlation gain occurs in all ten seeds. A8 does not significantly exceed A1 in ECI, so the result specifically isolates the benefit of saliency protection relative to an otherwise comparable random-lacuna control.

\subsection{Discussion}

The extended results refine, rather than replace, the main-paper analysis. Under ten matched seeds and validation-only selection, saliency-informed variants remain frequently selected, but neither
their recognition gains nor their optimal form are uniform across architectures and corpora. The findings therefore support saliency as a constraint space rather than as a universal erasure rule.
Protected-random, mixed, high-, and low-saliency variants are selected in different settings, while the random foreground-lacuna control remains strongest for one architecture--dataset pair. Thus, saliency-informed placement can improve validation-selected performance, but its recognition benefit is architecture- and corpus-dependent.

ECI adds information not contained in macro-F1. The significant ConvNeXt result on Corpus Litterarum shows that two similarly accurate conditions can differ in explanation stability when the only substantive change is whether salient regions are protected. At the same time, high ECI alone is insufficient: minimally altered inputs may preserve explanations without improving recognition.

Overall, the extended analysis supports the central main-paper claim that saliency can guide augmentation design, while qualifying the strength of the recognition claim: its benefits are
architecture-dependent, and explanation consistency may improve even when recognition differences are not statistically significant.

\bibliography{aaai2027}
\end{document}